# Stochastic Formulation of Causal Digital Twin – Kalman Filter Algorithm

Dr. PG Madhavan

Seattle, USA May 2021
pg@jininnovation.com

#Causality #Dynamics #DigitalTwin #Causaldigitaltwin #KalmanFilter #IoT #Causalgraph #Learning

**Abstract**: *We provide some basic and sensible definitions of different types of digital twins and recommendations on when and how to use them. Following up on our recent publication of the Learning Causal Digital Twin, this article reports on a stochastic formulation and solution of the problem. Structural Vector Autoregressive Model (SVAR) for Causal estimation is recast as a state-space model. Kalman filter (and smoother) is then employed to estimate causal factors in a system of connected machine bearings. The previous neural network algorithm and Kalman Smoother produced very similar results; however, Kalman Filter/Smoother may show better performance for noisy data from industrial IoT sources.*

Causality is the most fundamental connection in the Universe. The connection is the knowledge of cause and effect and chain of causes. Without knowing what causes what, one cannot develop "Explainable" ML (machine learning) nor can one identify ways to achieve production increase in an industrial plant.

In the past decade, Internet of Things (IoT) has become a familiar part of work and home life – "things" with sensors send data to the Internet and you can watch those things "live" on a screen. *While IoT is about "THINGS" (or assets), knowing cause-effect RELATIONSHIPS among "things" takes IoT capabilities to a higher level – this is provided by "CAUSAL digital twin" (CDT)*. Since relationships are always changing, we need a dynamical representation, much like a video and not just a snapshot.

Digital twin is a software representation of a physical object, be it a machine, a system of connected assets or a whole city. In a typical embodiment of a digital twin, measurements via Internet of Things (IoT) enliven the software representation. There are a few types of Digital Twins as shown in figure 1.

Display digital twin is the most common type already deployed in many IoT systems. They provide immediate visibility into to current state of "things". Forward digital twin – which is more commonly called "Simulation" digital twin - is generally a Physics-based computational digital twin most useful at the design stage of the product life cycle.

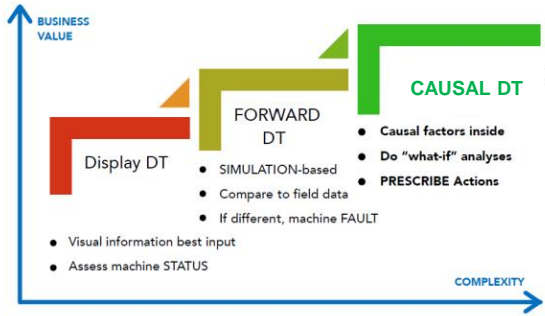

Figure 1. Digital twin types

CAUSAL digital twin (CDT) goes beyond these two types and captures the dynamics of interconnections among assets in terms of cause and effect relationships. The real purpose of developing and deploying such a digital twin is to understand the "parameters of the underlying system" so that we can *develop a CAUSAL understanding* of what the measurements are telling us about the system *("what is causing what?")*. Digital twin is where cause-effect determination can happen.

The ability to determine cause-effect relationships is lacking in IoT today; without this, "what-if" and "counterfactual" experimentations required for **prescriptive analytics** are not possible. *Prescriptive analytics tells a business what to do so that operations and production can improve – this requires causal relationships to be quantified beyond just correlations!*

Forward simulation may play a part in Display digital twin; simulation based on Causal Graphs are also part of CAUSAL digital twins. This commonality of simulation has led to much confusion. The table below addresses the main features of each digital twin and the best way to deploy them. To improve operational aspects, we have to focus on the DYNAMICS of a system. In the specific case of "machine dynamics", we are interested in the kinetics and kinematics of the machine and NOT the Statics or Structural aspects.

*In our recent companion article*, "[Evidence-based Prescriptive Analytics, CAUSAL Digital Twin and a Learning Estimation Algorithm](#)" (Madhavan, 2021), *we developed a Causal Digital Twin (CDT)*

| Digital Twin types → | Display DT | Forward DT | Causal DT |
|---|---|---|---|
| *Application:* | | | |
| Product design | | *** | |
| Maintenance | *** | | |
| Production/Ops | | | *** |
| *# of Assets:* | | | |
| Single Asset | ✓ | ✓ | |
| Interconnected Assets | | | ✓ |
| *System Type:* | | | |
| Structural (static) | * | *** | |
| Dynamics | | | *** |

solution and provided brief introductions to Causality basics, Causality in IOT and some Causality insights; we presented a solution that has a unique and novel neural network architecture incorporating forward Causal Graph simulation with backpropagation working across it.

*In the current article, we develop a Causal Digital Twin (CDT) solution that utilizes stochastic formulation. It lends itself to causal factor estimation using the powerful Kalman filter (and its variant called "Kalman Smoother").*

## Real Data Modeling: NASA Bearing data

We will develop Causal DT in a real-life setting using the popular NASA Prognostics Data Repository's bearing dataset. The data is from a run-to-failure test setup of bearings installed on a shaft. The arrangement is shown in figure 2.

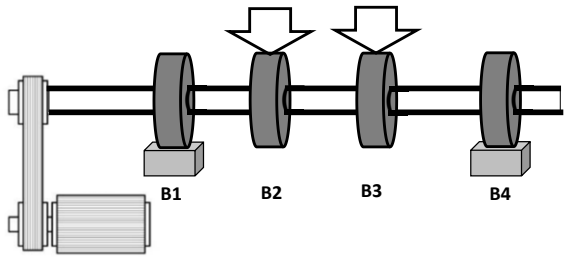

Figure 2. NASA Prognostic Data Repository Bearing data collection setup

Data were collected from Feb 12 to Feb 19, 2004 and the tests were run continuously to failure. Bearing 1 will be the target of our study – its outer race failed on Feb 19.

From the structure of the setup and some knowledge of machine dynamics, bearing operation and vibration analysis, we can come up with the Causal Graph shown in figure 3. For full details, see our companion article.

## State-space model

For the current study, we base our estimation method on Hyvarinen (2010), who developed the method for instantaneous and lagged causal effects using Structural Vector Autoregressive (SVAR) model.

General form of SVAR model:

$$\mathbf{y}_t = \mathbf{A}^0 \mathbf{y}_t + \sum_{m=1}^{M} \mathbf{A}^m \mathbf{y}_{t-m} + \mathbf{e}_t$$

All bolded quantities are vectors or matrices. $\mathbf{A}^0$ has zeros for its diagonal entries since self-causality does not exist. $\mathbf{y}_t$ are the measurements of the nodes of our Causal Graph. First term is Structural causality and the second term is Lagged causality. Casting SVAR in a state-space model:

$\underline{x}[n+1] = \mathbf{A}\,\underline{x}[n] + \underline{w}[n]$, where $\mathbf{A} = \mathbf{I}$

$\underline{y}[n] = \mathbf{H}[n]\,\underline{x}[n] + \underline{\varepsilon}[n]$

All bolded variables are matrices or vectors. The usual assumptions of process and measurement noise processes apply.

Here $\underline{y}[n] = [y_1\,y_2\,y_3\,y_4]^T$ are the vibration data from the 4 bearings. If $\mathbf{H}[n]$ is properly formed, State vector, $\underline{x}[n]$, will contain the causal factors that correspond to the causal graphs in figure 3.

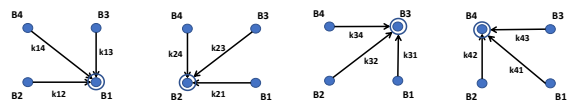

Figure 3. Causal graphs with Bearings 1, 2 3 & 4 as the Outcome Nodes of interest

Each outcome variable has 6 causal factors, 3 Instantaneous and 3 delayed (lag=1).

*With the state-space formulation, all 6*4=24 causal factors can be simultaneously estimated.*

## Kalman Filter / Smoother

Kalman filter is well understood and algorithm code snippets are available in public domain (example, Madhavan, 2016).

*We discard self-instantaneous and self-lagged effects.* Then any bearing data, y, can be written as –

$y_k[n] = \underline{Y}_k[n, n-1] \underline{X}_k[n]$

where $\underline{Y}_k[n, n-1] = [y_1[n] \ldots \cancel{y_k[n]} \ldots y_G[n]\ y_1[n-1] \ldots \cancel{y_k[n-1]} \ldots y_G[n-1]]$

and $\underline{X}_k[n] = [x_{k1}[n]\ x_{k2}[n]\ \ldots\ x_{kG}[n]\ x^L{}_{k1}[n]\ \ldots\ x^L{}_{kG}[n]]^T$

Therefore, we can write -

$$\begin{pmatrix} y_1[n] \\ \ldots \\ y_G[n] \end{pmatrix} = \begin{pmatrix} \underline{Y}_1[n, n-1] & \underline{Z} & \underline{Z} & \underline{Z} \\ & \ldots & & \\ \underline{Z} & \underline{Z} & \underline{Z} & \underline{Y}_G[n, n-1] \end{pmatrix} \begin{pmatrix} \underline{X}_1[n] \\ \ldots \\ \underline{X}_G[n] \end{pmatrix} = \underline{H}[n]\ \underline{x}[n]$$

**Z** is a zero vector of compatible dimensions. In NASA Bearing case, G=4.

**H[n]** is then a combination of $\underline{Y}_k[n, n-1]$ vectors and **Z** vectors.

Now, all the terms of the Kalman filter are defined and Kalman prediction and filter steps can be calculated for each data point as it arrives.

Note that Kalman initializations can be critical; the initial value of the covariance matrices for Process, Measurement noise processes and States have to be chosen carefully for numerical stability.

## Kalman Smoother Results

As we did in the [companion paper](#), we will focus on the dynamics of Bearing 1 which is the one that failed. Vibration time series is shown in figure 4.

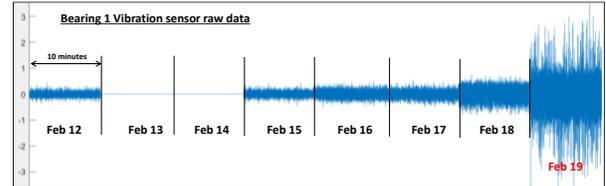

Figure 4. Time series of Bearing 1 vibration (NOT contiguous)

It turns out that the results of Kalman Smoother estimation is practically the same as the ones in the companion paper using neural networks. See figure 5.

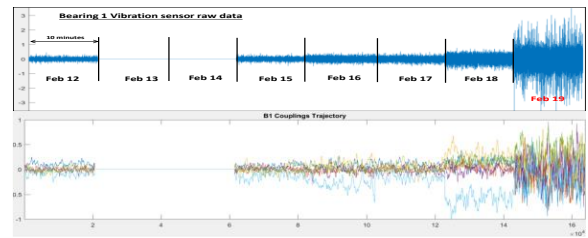

Figure 5. Kalman Smoother estimates of Causal Factor trajectories

Kalman Smoother provides a time-delayed estimate of causal factors. If real-time estimates are needed, Kalman filter can be used without much loss of accuracy; alternatively, a so-called "delayed" Smoother can be deployed where the backward pass of the Kalman updates for the Smoother is delayed by an acceptable duration.

## CONCLUSION

*We add to the foundational expository work on Causality theory applied to the DYNAMICS of connected system of assets by*

*adding a new and powerful stochastic algorithm for Causal Estimation*.

The previous algorithm in Madhavan (2021) was a novel, recurrent neural network with Causal Graph simulation incorporated into the network. The enhanced back-propagation algorithm to accommodate the simulation layer resulted in good performance.

Current work is purely stochastic and as such, for the type of noisy data we find in industrial IoT applications, Kalman solution may perform better. On the other hand, if we suspect strong nonlinear dynamics in the system of interconnected assets, neural network based solution may be more attractive.

Causal digital twin focuses on a system of connected assets and its dynamics; this causality focus of ours may be unique in IoT applications so far. *When connected assets in typical production environments are treated as a system, Prescriptive Analytics that result from Causal Digital Twin become directly useful in enhancing business outcomes such as increased production volume, better quality and reduced waste – all contributing to increased gross margin for the business*.

It should be noted that Causal Digital Twin is the next step in IoT system evolution! However, since monitoring single assets have already become prevalent, *NOW is the right time to conduct various proofs-of-concept in verticals such as Manufacturing, Utilities, Oil & Gas and Smart Buildings and be ready for rapid deployment within the next year of Causal digital twin solutions that will directly impact customer revenue!*

**About the author:**

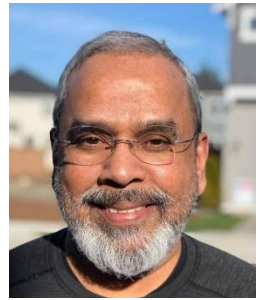

Dr. PG Madhavan launched his first IoT product at Rockwell Automation back in 2000 for predictive maintenance, an end-to-end solution including a display digital twin. Since then, he has been involved in the development of IoT technologies such as fault detection in jet engines at GE Aviation and causal digital twins to improve operational outcomes. His collected works in IoT is being published as a book, "Data Science for IoT Engineers" in June 2021. Rest of his career has been in industry spanning more major corporations (Microsoft, Lucent Bell Labs and NEC) and four startups (2 of which he founded and led as CEO).
https://www.linkedin.com/in/pgmad/